\documentclass[10pt,twocolumn,letterpaper]{article}

\usepackage{iccv} 

\usepackage{graphicx}
\usepackage{amsmath}
\usepackage{amssymb}
\usepackage{booktabs}
\usepackage{subcaption}
\usepackage{adjustbox}
\usepackage{colortbl}
\usepackage{booktabs}
\usepackage{hhline}
\usepackage{enumitem}
\usepackage{pifont} 
\usepackage{xcolor}
\usepackage{tabularray}
\usepackage{fancyhdr}
\usepackage[accsupp]{axessibility}

\newcommand{\colorcheck}[1]{\textcolor{#1}{\ding{51}}}
\newcommand{\colorcross}[1]{\textcolor{#1}{\ding{55}}}

\usepackage[pagebackref,breaklinks,colorlinks]{hyperref}

\usepackage[capitalize]{cleveref}
\crefname{section}{Sec.}{Secs.}
\Crefname{section}{Section}{Sections}
\Crefname{table}{Table}{Tables}
\crefname{table}{Tab.}{Tabs.}

\def\paperID{15554} 
\def\confName{ICCV}
\def\confYear{2025}

\begin{document}

\title{COME: Dual Structure-Semantic Learning with Collaborative MoE for Universal Lesion Detection Across Heterogeneous Ultrasound Datasets
}

\author{
\small
Lingyu Chen\textsuperscript{1,*},
Yawen Zeng\textsuperscript{2,*},
Yue Wang\textsuperscript{1},
Peng Wan\textsuperscript{1},
Guochen Ning\textsuperscript{3},
Hongen Liao\textsuperscript{3,4},
Daoqiang Zhang\textsuperscript{1},
Fang Chen\textsuperscript{4,\dag}\\
\small \textsuperscript{1}College of Artificial Intelligence, Nanjing University of Aeronautics and Astronautics\\
\small \textsuperscript{2}ByteDance Inc. \quad 
\textsuperscript{3}Tsinghua University \quad 
\textsuperscript{4}Shanghai Jiaotong University\\
\small \texttt{lingyucher@163.com, yawenzeng11@gmail.com, chenfang\_bme@163.com}
}


\maketitle

\footnotetext[1]{*Equal contribution.}
\footnotetext[2]{\dag Corresponding author.}


\begin{abstract}
Conventional single-dataset training often fails with new data distributions, especially in ultrasound (US) image analysis due to limited data, acoustic shadows, and speckle noise.
Therefore, constructing a universal framework for multi-heterogeneous US datasets is imperative. However, a key challenge arises: how to effectively mitigate inter-dataset interference while preserving dataset-specific discriminative features for robust downstream task? 
Previous approaches utilize either a single source-specific decoder or a domain adaptation strategy, but these methods experienced a decline in performance when applied to other domains. Considering this, we propose a Universal Collaborative Mixture of Heterogeneous Source-Specific Experts (COME). Specifically, COME establishes dual structure-semantic shared experts that create a universal representation space and then collaborate with source-specific experts to extract discriminative features through providing complementary features. This design enables robust generalization by leveraging cross-datasets experience distributions and providing universal US priors for small-batch or unseen data scenarios. Extensive experiments under three evaluation modes (single-dataset, intra-organ, and inter-organ integration datasets) demonstrate COME's superiority, achieving significant mean AP improvements over state-of-the-art methods. Our project is available at: \url{https://universalcome.github.io/UniversalCOME/}.
\end{abstract}

\section{Introduction}
\label{sec:intro}

\begin{figure}[t]
  \centering
   \includegraphics[width=\linewidth]{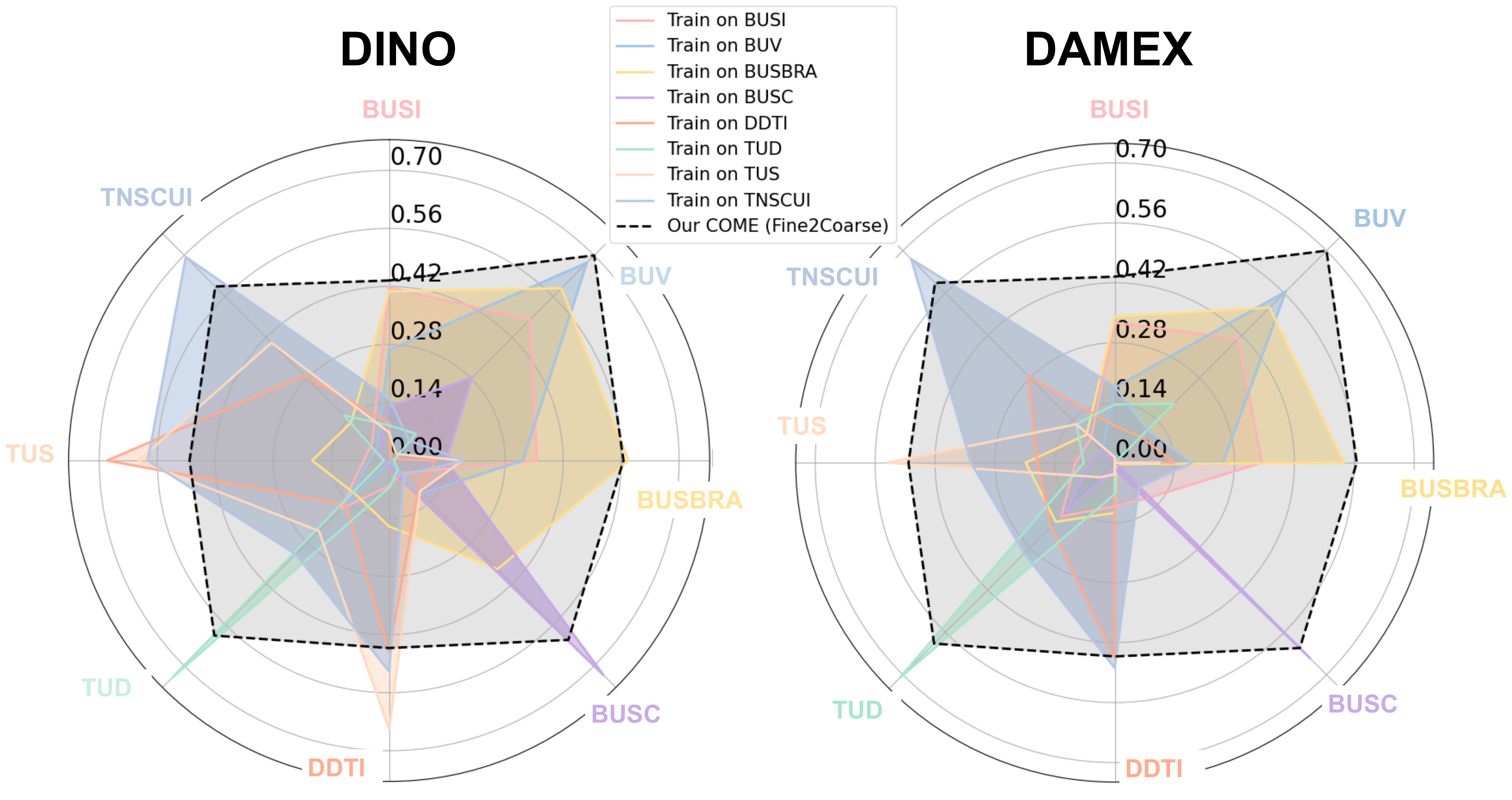}
   \vspace{-0.5cm}
   \caption{
   DINO \cite{zhang2022dino} and DAMEX \cite{jain2023damex} are trained on 8 datasets separately and then tested zero-shot on 7 others. In contrast, our COME achieves top performance on all 8 datasets by training on a combined heterogeneous US dataset.
   }
   \label{fig:radar}
   \vspace{-0.6cm}
\end{figure}

Modern automated ultrasound (US) analytics, with their advanced understanding of regions of interest, have reached clinical grade performance in detecting lesions, and segmentation \cite{chen2024versatile, han2022privacy, li2023dsmt, pei2023multi, chen2023deep, chen2020tissue}. However, despite improving diagnostic processes and treatment plans, their clinical application is limited because prevailing methods \cite{liu2024grounding, wang2025yolov10, chen2024sonographers, chen2024thinking} anchor training to single datasets, rendering them brittle under cross-dataset generalization.
For instance, as shown in \cref{fig:radar}, the DINO model trained on the TUD dataset \footnote{https://github.com/NEU-LX/TUD-Datebase.} performs poorly with other datasets, all scoring below 0.2, highlighting a gap between algorithmic complexity and real clinical reliability!

The limitations of single-dataset training paradigms and the scarcity of clinician-dependent high-quality US datasets \cite{chen2024ultrasound, wang2024uni}, \textit{motivate our universal framework for learning multiple heterogeneous datasets.} Firstly, we employ two strategies for the construction of a universal benchmark dataset, which originate from diverse imaging devices and institutions: \textbf{intra-organ consolidation} (aggregating representations of specific anatomical structures) and \textbf{inter-organ integration} (discovering latent correlations across distinct anatomies), exemplified by BUSI (633 cases \footnote{https://www.kaggle.com/datasets/sabahesaraki/breast-ultrasound-images-dataset/data.}) and BUSBRA (875 cases \cite{gomez2024bus}), as demonstrated in Fig.1 of Appendix. Particularly, it leverages clinician-curated knowledge and enriches feature patterns with growing public datasets. As stated in \cite{chen2024multi, wu2023cross}, cross-dataset operation enhances feature interdependencies while maintaining unique patterns. Our universal framework ensures consistent structure, semantic stability, and effective feature extraction, providing a reliable basis for US analysis.

In fact, developing universal models through the integration of multiple heterogeneous US datasets faces challenges: (1) inherent heterogeneity (including artifact distributions and echogenicity variations) from diverse source inducing domain shifts \cite{renaud2024sources} and (2) imbalanced number of images exist across datasets. This compounded heterogeneity underscores the demand for frameworks that moderate variations while preserving consistent semantics, a crucial unmet need in generalizable US analytics. As shown in \cref{table.1}, current methodologies involving multiple medical heterogeneous datasets incorporate unique architectural or training designs. For example, MVFA \cite{huang2024adapting} requires constructing separate encoders for different sources; models such as SLDG \cite{wu2023iterative} and Dataseg \cite{gu2023dataseg} incorporate task-specific heads to avoid performance drops when encountering new data; approaches like MedUniSeg \cite{ye2024meduniseg} and MapSeg \cite{zhang2024mapseg} need prior knowledge of test domain information; and MiDSS \cite{ma2024constructing} demands additional fine-tuning when applied to the target dataset. However, these methods contradict the goal of source-agnostic generalization: strong domain dependency, constrained cross-modal design, and insufficient anatomical representation hinder effective balancing of cross-dataset interference suppression and intra-dataset discriminative feature preservation.

\begin{table}
\centering
\caption{Compared to others, COME offers architectural advantages with fewer constraints. More \colorcheck{green} indicate greater restrictions.}
\vspace{-0.3cm}
\label{table.1}
\resizebox{0.48\textwidth}{!}{
\begin{tabular}{l||cccc} 
\toprule
\textbf{Benchmark} & \begin{tabular}[c]{@{}c@{}}\textbf{Separate }\\\textbf{Encoders}\end{tabular} & \begin{tabular}[c]{@{}c@{}}\textbf{Task-specific }\\\textbf{Heads}\end{tabular} & \textbf{Test Source} & \textbf{Tuning}  \\ 
\toprule
SLDG \cite{wu2023iterative}         & \colorcross{red}                                       & \colorcheck{green}                                         & \colorcross{red}                    & \colorcross{red}               \\
MedUniSeg \cite{ye2024meduniseg}    & \colorcross{red}                                       & \colorcross{red}                                         & \colorcheck{green}                    & \colorcross{red}               \\
MVFA \cite{huang2024adapting}        & \colorcheck{green}                                       & \colorcross{red}                                         & \colorcheck{green}                    & \colorcross{red}               \\
Dataseg \cite{gu2023dataseg}      & \colorcross{red}                                       & \colorcheck{green}                                         & \colorcross{red}                    & \colorcross{red}               \\
MapSeg \cite{zhang2024mapseg}      & \colorcross{red}                                       & \colorcross{red}                                         & \colorcheck{green}                    & \colorcheck{green}               \\
MiDSS \cite{ma2024constructing}        & \colorcross{red}                                      & \colorcross{red}                                         & \colorcross{red}                    & \colorcheck{green}               \\
\toprule
COME (Ours)        & \colorcross{red}                                      & \colorcross{red}                                         & \colorcross{red}                    & \colorcross{red}   \\
\toprule
\end{tabular}}
\vspace{-0.6cm}
\end{table}

To address these issues, we propose the universal \textbf{\underline{Co}llaborative \underline{M}ixture of Heterogeneous \underline{E}xperts (COME)}, evaluated on US lesion detection tasks. Specifically, considering the common features (e.g., structural consistency and high-level semantic invariance) in multiple heterogeneous datasets, we introduce two complementary shared experts: \underline{St}ructure Shared \underline{E}xpert (\textbf{STE}) provides structural priors through a feature extractor pre-trained on multi-organ US corpus \cite{jiao2024usfm} using masked image modeling (MIM) in spatial and frequency domains; \underline{Se}mantic Shared  \underline{E}xpert (\textbf{SEE}) aligns semantic features via MedCLIP \cite{koleilat2024medclip} pre-trained with multi-modal US images-text pairs \cite{li2024ultrasound} for complementary representation. Simultaneously, COME applies the proposed Fine2Coarse or Multi-Step clustering to tokens after multi-head attention, routes to Mixture of Heterogeneous \underline{S}ource-\underline{S}pecific \underline{E}xperts (\textbf{S²E}), then concatenate with dual shared universal features for downstream networks. COME mitigates multi-dataset interference while utilizing S²E to extract discriminative features through collaborative training. Extensive validation on eight diverse US datasets, covering both breast and thyroid organs, highlights COME’s state-of-the-art performance and superior capabilities. Our contributions can be summarized:

\begin{itemize}[leftmargin=*]
 \item{\textbf{Universal Framework.} We design a practical yet challenging task to implement a universal framework on multiple heterogeneous US datasets and validate it with lesion detection tasks.}
 \item{\textbf{COME Model.} Our COME is an MoE structure that collaborates with two shared experts, which captures shared semantic patterns and retains unique dataset features to enrich latent representation.}
 \item{\textbf{Experimental Performance.} Empirical results demonstrate that our method significantly improves detection performance for multiple heterogeneous US datasets.}
\end{itemize}


\section{Related Work}

\begin{figure*}[t]
  \centering
   \includegraphics[width=\linewidth]{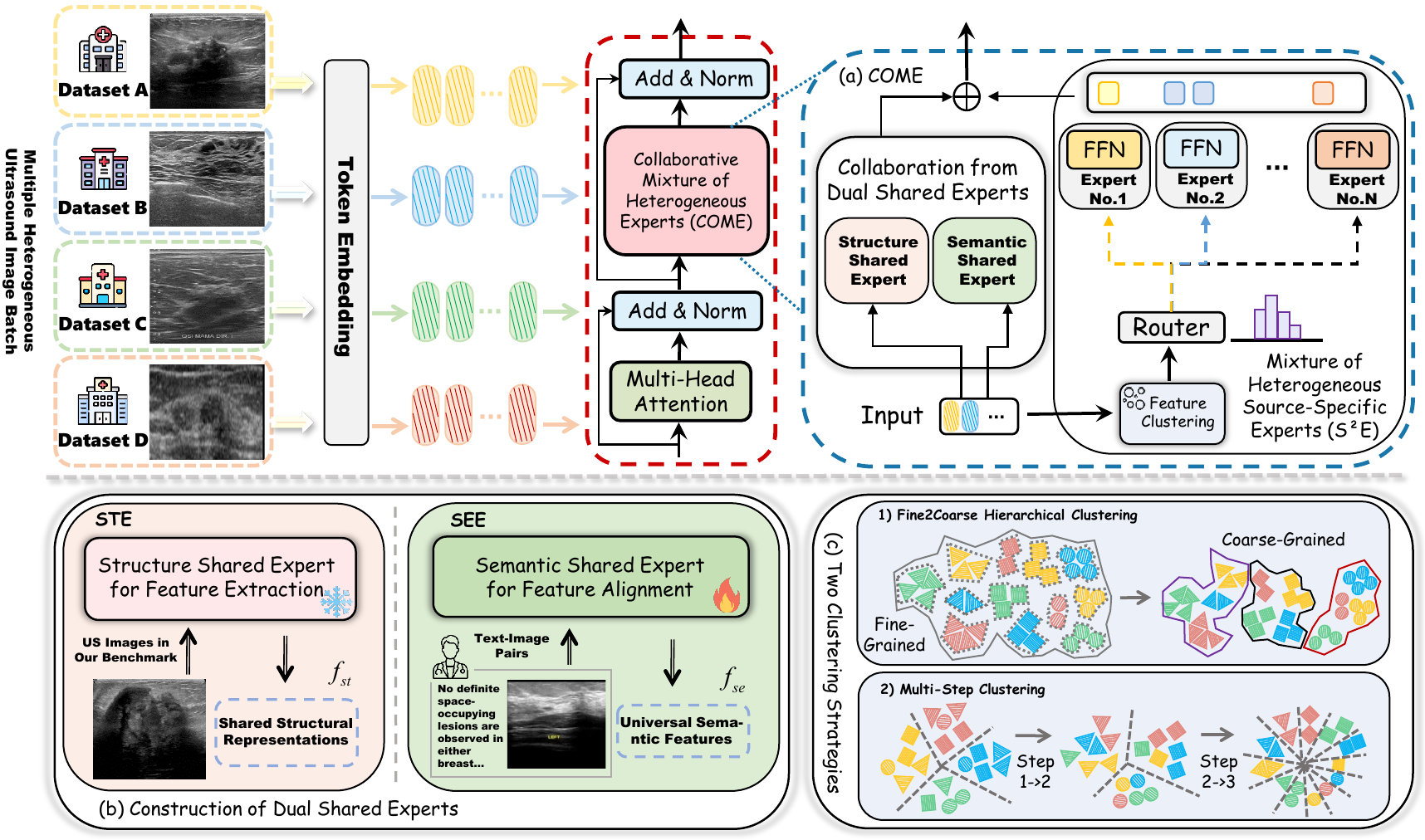}
   \vspace{-0.5cm}
   \caption{\textbf{(a) Overview} of COME. \textbf{(b) Two shared experts} are constructed based on structure-semantic learning for the latent universal feature space. \textbf{(c) Two clustering strategies} in the S²E route tokens from the same dataset to specific experts, enabling expert specialization.}
   \label{fig:model}
   \vspace{-0.4cm}
\end{figure*}

\textbf{Training on Multiple Heterogeneous Medical Datasets.} 1) \textit{In terms of architectural design}, existing universal models still rely on domain-specific knowledge and struggle to balance parameter complexity with generalizability \cite{xiong2024one}. They capture global cues but miss fine-grained cross-dataset details \cite{awais2025foundation}, often requiring retraining that heightens domain dependence and undermines generalizability. Source-specific detectors \cite{wu2023iterative} with isolated encoders suffer performance degradation when directly fusing task-specific heads. Dataseg \cite{gu2023dataseg} uses a multi-scale architecture with shared parameters but needs domain-specific heads. Residual adapter-based model \cite{huang2024adapting} depends on source-dependent feature banks for inference alignment, introducing computational latency and failing in unseen source scenarios. MedUniSeg \cite{ye2024meduniseg} achieves semantic decoupling in shallow layers via learnable source-specific prompts combined with task-general prompts, but requires source information during inference. 
Despite competitive performance, these methods ignore common features across datasets and add complexity with specific components or retraining.


2) \textit{Regarding data modalities}, we have observed a growing number of studies focusing on mining datasets from different organs. MOFO \cite{chen2024multi} achieves anatomical consistency with multi-organ perception and prompt templates, but is still prone to capture nonlinear heterogeneous feature dependencies and sensitivity to cross-anatomical commonality. GSViT \cite{schmidgall2024general} uses surgical videos to model tissue deformation, yet its reliance on controlled conditions limits clinical use. SegAnyPath \cite{wang2024seganypath} employs a multi-scale proxy module to represent organ features but requires explicit pathological annotations. Lvm-med \cite{mh2023lvm} enforces structural constraints with similarity metrics, assuming organ-specific references during testing. Though multi-organ analysis has improved, these models still grapple with the complexity of US imaging due to neglecting its characteristics.

\noindent\textbf{Foundation Models for Medical Heterogeneous Data.} Universal US models (USFM \cite{jiao2024usfm}, SAMMedUS \cite{tian2025sam}, UltraSAM \cite{meyer2024ultrasam}) lack domain-specific knowledge, causing suboptimal detection performance on irregular lesion \cite{huang2024adapting}, and requiring further fine-tuning \cite{jain2023damex}. Specifically, USFM \cite{jiao2024usfm} uses dual-domain masking (spatial-frequency) for self-supervised pre-training, aiding robust semantic learning from US-specific patterns. Endo-FM \cite{wang2023foundation} captures long-range dependencies in polyp videos, enhancing coherence with global-context attention. PathoDuet \cite{hua2024pathoduet} integrates histopathological patterns during vision-language pre-training.
Fine-tuning on single datasets improves performance, but struggles across different datasets due to domain shifts and modality variations.

\noindent\textbf{Mixture-of-Experts.} The MoE framework uses expert networks and a dynamic router to process heterogeneous data. Each expert focuses on specific patterns (e.g., acoustic artifacts, organ morphology), while the router optimizes feature fusion with attention weights. Traditional MoE methods \cite{chen2023adamv, yu2024boosting, xue2023raphael}, have issues with biased routing, leading to under-utilized experts and load imbalance. Recent approaches have trade-offs: DeepseekMoE \cite{dai2024deepseekmoe} isolates shared experts but uses fixed sharing ratios; Zhang et al. \cite{zhang2024efficient} implicitly instantiates single shared expert via modulation matrices, sacrificing feature disentanglement; HyperMoE\cite{zhao2024hypermoe} uses hypernetworks for cross-domain adaptation. Inspired by these developments, we suggest that shared experts are crucial for cross-dataset feature complementarity.


\section{Methodology}

\subsection{Overall Framework}

We propose the \underline{\textbf{Co}}llaborative \underline{\textbf{M}}ixture of Heterogeneous \underline{\textbf{E}}xperts (COME), a novel framework for lesion detection across multiple heterogeneous US datasets that synergistically integrates dual structural-semantic priors with intra-dataset specific feature disentanglement. 

As illustrated in \cref{fig:model}, the framework extends the DINO-ViT backbone \cite{zhang2022dino, jain2023damex}, comprising of an encoder to undergo unified patch embedding and a decoder for downstream inference, both of which have 6 Transformer layers, and we innovatively replace standard FFN layers with COME. Each COME consists of a \textbf{St}ructure Shared \textbf{E}xpert (STE) to learn dataset-agnostic imaging and structural consistency, a \textbf{Se}mantic Shared \textbf{E}xpert (SEE) to encode high-level invariant anatomical semantics, and a heterogeneous \textbf{S}ource-\textbf{S}pecific \textbf{E}xpert module (S²E) for discriminative representations. STE and SEE provide dual shared prior knowledge to augment S²E's capacity for feature enrichment during single-source information mining. 

Formally, given a set of multiple US datasets $\{{{USd}_{1}},{{USd}_{2}},...,{{USd}_{M}}\}$, where denotes the i-th source dataset containing images $\{x_{1}^{i},x_{2}^{i},...,x_{n}^{i}\}$, $x\in {{\mathbb{R}}^{H\times W\times 3}}, n=\left| {{USd}_{i}} \right|$, model processes inputs $X\in {{USd}_{i}}$ to predict target $Y$ (bounding box coordinates, category labels, confidence scores). 

In the following sections, we will present the design motivations and details of each module (i.e., STE, SEE and S²E) individually.

\subsection{Collaborative Mixture of Heterogeneous Source-Specific Experts}

Following the structural design of MoE \cite{dai2024deepseekmoe,zhao2024hypermoe}, the overview of COME can be defined as:
\begin{equation}
    {{F}^{E}}={{E}_{st}}(X)+{{E}_{se}}(X)+{{E}_{{{S}^{2}}}}(Token(X)).
\end{equation}
In this formulation, the input token $Token(X)$ is fed to the S²E module and router network, which captures the source-dependent representation. Simultaneously, we design two shared experts, STE ${{E}_{st}}$ and SEE ${{E}_{se}}$, to extract semantic features. Finally, the aggregate result is ${{F}^{E}}$.


\noindent\textbf{3.2.1 Structure Shared Expert (STE).} We integrate a pretrained USFM \cite{jiao2024usfm} as the foundational component, which employs MIM with dual masking in spatial and frequency domains to extract robust low/mid-level anatomical features. Specifically, spatial-domain masking enhances awareness of stable structural patterns, while frequency-domain learnable masks refine implicit grayscale distributions and texture variations. Given USFM's extensive pretraining on large US data with noises and heterogeneity, we freeze its parameters as a fixed feature extractor ${{E}_{st}}(\cdot )$ during COME training to preserve structural representations ${{f}_{st}}\in {{\mathbb{R}}^{pH\times pW\times C}}$. This design not only achieves effective hierarchical feature extraction, but also mitigates gradient conflict during optimization.

\noindent\textbf{3.2.2 Semantic Shared Expert (SEE).} Inspired by the gains from pre-training, we aim to use multimodal US models to enhance COME with universal semantic features via cross-modal alignment. Specifically, firstly, we fine-tune MedClip-SAMv2 \cite{koleilat2024medclip} using  $N$ multimodal US image-text pairs $\{I_{i}^{US},T_{i}^{US}\}_{i=1}^{N}$ of the thyroid and breast. It integrates BiomedCLIP \cite{zhang2023biomedclip} and SAM \cite{kirillov2023segment} to achieve pixel-text semantic alignment. Secondly, given the lack of corresponding text data in our benchmark, inspired by \cite{koleilat2024medclip, huang2024adapting}, we construct the structured text corpus ${USt}$ (\textit{Diagnosis for \textcolor{orange}{\textit{\{Organ\}}} Ultrasound Image: \textcolor{orange}{\textit{\{Benign/Malignant/Omit\}}}}) based on lesion attributes with fixed prompts, using diagnosis labels from datasets and forming paired training materials $\{{{USd}_{i}}, {{USt}_{i}}\}_{i=1}^{M}$. Semantic-guided universal features ${{f}_{gse}}\in {{\mathbb{R}}^{pH'\times pW'\times C'}}$ are then extracted from input images $X$. During testing, we just use "Diagnosis for Ultrasound Image". This process maximizes the use of limited textual resources and offers rich prior knowledge, improving adaptability to small-scale clinical datasets.

\noindent\textbf{3.2.3 Mixture of Heterogeneous Source-Specific Experts (S²E).} This module ${{E}_{{{S}^{2}}}}$ comprises ${M}'$ expert networks and a gating router ${{G}_{{{S}^{2}}}}$, designed to extract discriminative features from individual dataset. MoE excels at dynamically encoding features for each sample while integrating multiple experts to enhance representation diversity. Expert Networks are denoted as $\{{{E}_{1}},{{E}_{2}},...,{{E}_{{{M}'}}}\}$, ${M'}>{M}$. We introduce an innovative strategy where a group of experts focuses on patterns in a specific dataset to avoid loss of sensitivity due to multi-source complexity. Ideally, each expert should specialize in single dataset features; however, dataset variations may introduce noise, affecting specialization. To solve this, we cluster before feature routing using two strategies:

\textbf{1) Fine2Coarse Hierarchical Clustering} is a two-phase hierarchical clustering designed to balance local specificity and global consistency. The motivation for this clustering arises from the need to avoid the pitfalls of overly coarse groupings, which can overlook important local patterns, and excessively fine-grained partitions, which can lead to unnecessary fragmentation of a single source. Specifically, we first employ K-means \cite{ahmed2020k} for \textbf{fine-grained clustering}:
\begin{equation}
    A\leftarrow K-means(Multi-Head Att(Token(X)))
\end{equation}

Let $A$ represent the learned cluster centroids, where the number of fine-grained clusters, $m=\left| A \right|$ is assigned an elevated value to ensure each centroid optimally encapsulates semantically cohesive representations of its constituent features. This operation facilitates granular discernment of latent patterns, particularly critical for capturing nuanced inter-source variations. 
Thereafter, we implement hierarchical clustering refinement: The preliminary fine-grained clusters undergo secondary aggregation through \textbf{coarse-grained clustering}:
\begin{equation}
    B\leftarrow K-means(A)
\end{equation}

Ultimately, we extract the corresponding feature representations ${{F}_{c}}$ for each coarse-grained cluster centroid $c\in B$, forming the cluster feature set $F={{\cup }_{c\in C}}{{F}_{c}}$. We then merge the original features with the clustered features to construct the input representation ${F}'$ for the routing network, followed by dimension reduction $DR( \cdot )$:
\begin{equation}
    {F}'=DR(Cat(Multi-Head Att(Token(X)), F))
\end{equation}

\textbf{2) Multi-Step Clustering} is designed to address the challenges of attribute entanglement and feature coupling in heterogeneous data. Specifically, the clustering results from the previous step are used as priors in the current clustering process, thereby facilitating semantic continuity across steps and adaptively suppressing anomalous clusters.


\noindent\textbf{3.2.4 Router Network.} The dynamic routing network ${{G}_{{{S}^{2}}}}: {{\mathbb{R}}^{D}}\to {{\mathbb{R}}^{{{M}'}}}$, implemented as a lightweight linear transformation layer, maps clustered features ${F}'\in {{\mathbb{R}}^{D}}$ to expert activation scores with the learnable matrix ${{W}_{g}}\in {{\mathbb{R}}^{{M}'\times D}}$. For the j-th expert, the probability ${{g}_{j}}$ is as follows:
\begin{equation}
    {{g}_{j}}=\frac{exp({{W}_{g,j}}{F}'+{{b}_{g,j}})}{\sum\nolimits_{n=1}^{{{M}'}}{exp({{W}_{g,n}}{F}'+{{b}_{g,n}})}}
\end{equation}

Ultimately, the temperature-scaled Top-K experts are utilized to route the tokens. Final features are computed as a weighted combination of the processed tokens and the activation probabilities:
\begin{equation}
    {{f}_{{{S}^{2}}}}={{E}_{{{S}^{2}}}}(Token(X))=\sum\limits_{j\in Top-k}{{{g}_{j}}{{E}_{j}}({F}')}
\end{equation}

\subsection{Collaborative Source-Specific Loss Function}

To address expert learning bias and load imbalance during training, we introduce dual-constrained joint optimization objectives. On the one hand, our source-specific traceability loss $\mathcal{L}_{TB}$ establishes interpretable expert-source mapping by enhancing intra-source routing consistency and inter-source expert distinction. This aims to assign domain-specialized experts to mine source-specific features, ensuring homogeneous inputs from the same source are routed to identical experts. Optimized through cross-entropy supervision, this loss promotes expert specialization while maintaining balanced utilization:
\begin{equation}
    \mathcal{L}_{TB}=-\sum_{j=1}^{|E|}\mathbf{1}(d_j=j)\log(g_j)
\end{equation}

Here, ${{d}_{j}}$ represents the expert label of the source dataset $US{{d}_{j}}$ for input tokens, and ${{g}_{j}}$ denotes the probability of the selected expert.

On the other hand, practical training often deviate from ideal conditions: input tokens from same dataset may be distributed across multiple experts rather than concentrated in one, and the Top-K routing mechanism induces expert activation sparsity. Thus, we also add a load balancing loss ${{\mathcal{L}}_{balance}}$ comprising importance loss ${{\mathcal{L}}_{IP}}$ and load loss ${{\mathcal{L}}_{Load}}$. We first compute the importance $I{{P}_{j}}=\sum\limits_{x\in Token(X)}{{{g}_{j}}}$ of each expert ${{E}_{j}}$, then derive the load to represent the distribution across experts ${{L}_{j}}=\sum\limits_{x\in Token(X)}{S({{g}_{j}})}$, and finally calculate them using the square of the coefficient of variation:
\begin{equation}
    {{\mathcal{L}}_{IP}}=CV{{(IP)}^{2}}
\end{equation}
\begin{equation}
    {{\mathcal{L}}_{Load}}=CV{{(L)}^{2}}
\end{equation}

Here, $S(\cdot )$ is the CDF of the standard normal distribution $\mathcal{N}$, and $CV(\cdot )$ is Coefficient of Variation. The sum of the two terms constitutes the ${{\mathcal{L}}_{balance}}$, and other losses in our universal model follow the DINO.


\begin{table*}
\caption{Lesion detection results on our benchmark, divided into top, middle, and bottom sections based on training data paradigms. We report mean AP score on each dataset. Among them, \textbf{bold} and \underline{underline} indicate optimal and sub-optimal performance for three paradigms.}
\label{table.2}
\centering
\setlength{\extrarowheight}{0pt}
\addtolength{\extrarowheight}{\aboverulesep}
\addtolength{\extrarowheight}{\belowrulesep}
\setlength{\aboverulesep}{0pt}
\setlength{\belowrulesep}{0pt}
\resizebox{17.5cm}{!}{
\begin{tabular}{c||c||cccccccc||c} 
\toprule
Method                                                   & Paradigms   & BUSI                                                & BUV                                                 & BUSBRA                                              & BUSC                                                                      & DDTI                                                & TUD                                                 & TUS                                                 & TNSCUI                                              & Mean                                                 \\ 
\hhline{=::=::========::=}
Open-GroundingDINO \cite{liu2024grounding}                                     & Single      & 0.2495                                              & 0.3649                                              & 0.3025                                              & 0.3666                                                                    & {\cellcolor[rgb]{1,0.925,0.518}}\textbf{0.5391}     & 0.7255                                              & {\cellcolor[rgb]{1,0.925,0.518}}\textbf{0.6112}     & {\cellcolor[rgb]{1,0.925,0.518}}\textbf{0.6986}     & 0.4822                                               \\
YOLOv10 \cite{wang2025yolov10}                                                 & Single      & 0.3161                                              & \underline{0.6216}                                              & \underline{0.5627}                                              & \underline{0.7052}                                                                    & 0.3640                                              & {\cellcolor[rgb]{1,0.925,0.518}}\textbf{0.7450}     & 0.4210                                              & 0.6907                                              & 0.5532                                               \\
DINO \cite{zhang2022dino}                                                     & Single      & {\cellcolor[rgb]{1,0.925,0.518}}\textbf{0.4180}     & {\cellcolor[rgb]{1,0.925,0.518}}\textbf{0.6754}     & {\cellcolor[rgb]{1,0.925,0.518}}\textbf{0.5780}     & {\cellcolor[rgb]{1,0.925,0.518}}\textbf{0.7313}                           & \underline{0.4609}                                              & \underline{0.7373}                                              & \underline{0.6047}                                              & \underline{0.6945}                                              & {\cellcolor[rgb]{1,0.925,0.518}}\textbf{0.6125}      \\
DINO-MoE                                                 & Single      & \underline{0.3388}                                              & 0.5510                                              & 0.5376                                              & 0.6706                                                                    & 0.4189                                              & 0.7088                                              & 0.5248                                              & 0.6669                                              & 0.5521                                               \\
DAMEX \cite{jain2023damex}                                                   & Single      & 0.3270                                              & 0.5653                                              & 0.5358                                              & 0.6458                                                                    & 0.4572                                              & 0.7101                                              & 0.5313                                              & 0.6727                                              & \underline{0.5556}                                               \\ 
\toprule
CerberusDet \cite{tolstykh2024cerberusdet}                                             & Intra-Organ & 0.3470                                              & 0.6980                                              & 0.6300                                              & \multicolumn{1}{c||}{0.5770}                                              & 0.5180                                              & 0.6780                                              & {\cellcolor[rgb]{0.557,0.737,0.561}}\textbf{0.6370}                                              & 0.6830                                              & 0.5960                                               \\
DINO                                                     & Intra-Organ & 0.3509                                              & 0.6399                                              & 0.5193                                              & \multicolumn{1}{c||}{{\cellcolor[rgb]{0.557,0.737,0.561}}\textbf{0.7112}}                                              & 0.4487                                              & 0.6899                                              & 0.5666                                              & 0.6597                                              & 0.5732                                               \\
DINO-MoE                                                 & Intra-Organ & 0.3269                                              & 0.6795                                              & 0.5419                                              & \multicolumn{1}{c||}{0.6710}                                               & 0.4898                                              & 0.7001                                              & 0.5801                                              & 0.6804                                              & 0.5837                                               \\
DAMEX                                                    & Intra-Organ & 0.3591                                              & 0.6990                                              & 0.5404                                              & \multicolumn{1}{c||}{0.6693}                                              & 0.5126                                              & 0.7000                                              & 0.5850                                              & \underline{0.6881}                                              & 0.5941                                               \\
\rowcolor[rgb]{0.824,0.827,0.827} Our COME (Multi-Step)  & Intra-Organ & \underline{0.4883}                                              & {\cellcolor[rgb]{0.557,0.737,0.561}}\textbf{0.7995} & {\cellcolor[rgb]{0.557,0.737,0.561}}\textbf{0.6748} & \multicolumn{1}{c||}{\underline{0.6879}} & {\cellcolor[rgb]{0.557,0.737,0.561}}\textbf{0.5481} & \underline{0.7032}                                              & 0.5900                                              & 0.6875                                              & {\cellcolor[rgb]{0.557,0.737,0.561}}\textbf{0.6474}  \\
\rowcolor[rgb]{0.824,0.827,0.827} Our COME (Fine2Coarse) & Intra-Organ & {\cellcolor[rgb]{0.557,0.737,0.561}}\textbf{0.4885} & \underline{0.7876}                                              & \underline{0.6689}                                              & \multicolumn{1}{c||}{0.6434}                                              & \underline{0.5443}                                              & {\cellcolor[rgb]{0.557,0.737,0.561}}\textbf{0.7039} & \underline{0.5971} & {\cellcolor[rgb]{0.557,0.737,0.561}}\textbf{0.6922} & \underline{0.6407}                                               \\ 
\toprule
CerberusDet                                              & Inter-Organ & 0.4570                                              & \underline{0.7940}                                              & 0.6490                                              & 0.6300                                                                    & \underline{0.5500}                                              & 0.6600                                              & {\cellcolor[rgb]{1,0.851,0.824}}\textbf{0.6130}     & 0.6620                                              & 0.6268                                               \\
DINO                                                     & Inter-Organ & 0.3825                                              & 0.6597                                              & 0.5395                                              & 0.7088                                                                    & 0.4583                                              & 0.6912                                              & 0.5475                                              & 0.6461                                              & 0.5792                                               \\
DINO-MoE                                                 & Inter-Organ & 0.4003                                              & 0.6838                                              & 0.5560                                              & 0.7227                                                                    & 0.5118                                              & {\cellcolor[rgb]{1,0.851,0.824}}\textbf{0.7103}     & 0.5799                                              & 0.6973                                              & 0.6077                                               \\
DAMEX                                                    & Inter-Organ & 0.4098                                              & 0.7007                                              & 0.5731                                              & \underline{0.7260}                                                                    & 0.5214                                              & \underline{0.7092}                                              & \underline{0.5994}                                              & 0.6942                                              & 0.6167                                               \\
\rowcolor[rgb]{0.824,0.827,0.827} Our COME (Multi-Step)  & Inter-Organ & \underline{0.4958}                                              & 0.7859                                              & {\cellcolor[rgb]{1,0.851,0.824}}\textbf{0.6912}     & 0.7191                                                                    & {\cellcolor[rgb]{1,0.851,0.824}}\textbf{0.5503}     & 0.7025                                              & 0.5972                                              & \underline{0.7049}                                              & \underline{0.6558                }                               \\
\rowcolor[rgb]{0.824,0.827,0.827} Our COME (Fine2Coarse) & Inter-Organ & {\cellcolor[rgb]{1,0.851,0.824}}\textbf{0.5159}     & {\cellcolor[rgb]{1,0.851,0.824}}\textbf{0.8313}     & \underline{0.6719}                                              & {\cellcolor[rgb]{1,0.851,0.824}}\textbf{0.7266}                           & 0.5371                                              & 0.7091                                              & 0.5725                                              & {\cellcolor[rgb]{1,0.851,0.824}}\textbf{0.7052}     & {\cellcolor[rgb]{1,0.851,0.824}}\textbf{0.6587}      \\
\toprule
\end{tabular}}
\end{table*}

\section{Experiments}
\label{sec:experiments}

\subsection{Dataset and Setting}

\begin{figure*}[!t]
  \centering
   \includegraphics[width=0.85\linewidth]{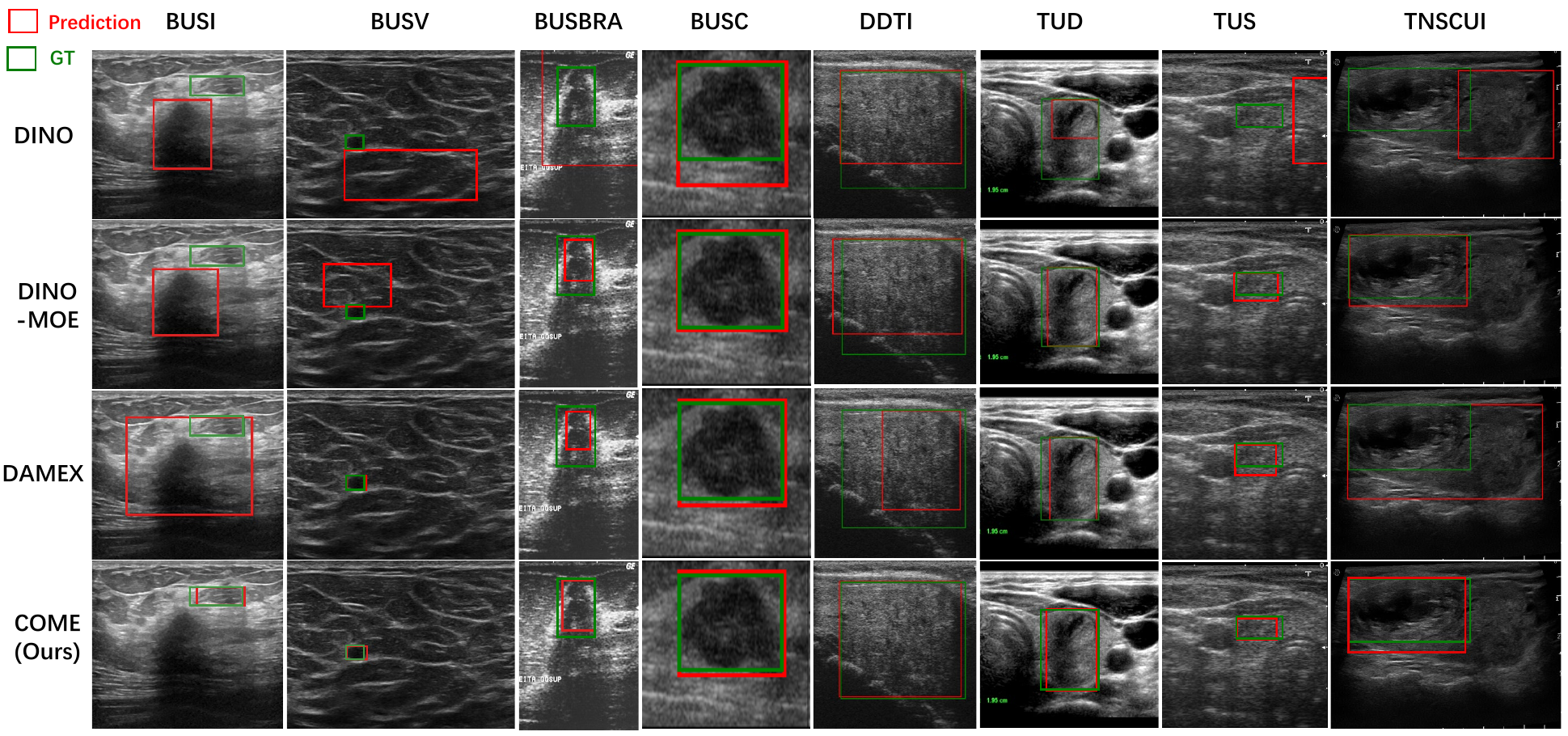}
   \caption{
   We visualize lesion detection on the inter-organ integration benchmark, comparing the non-MoE model DINO, the MoE-based DINO-MoE and DAMEX, and ours. COME demonstrates strong performance across all heterogeneous datasets.}
   \label{fig:detection}
   \vspace{-0.3cm}
\end{figure*}

\begin{figure*}[!t]
  \centering
   \includegraphics[width=0.9\linewidth, height=6.5cm, keepaspectratio=false]{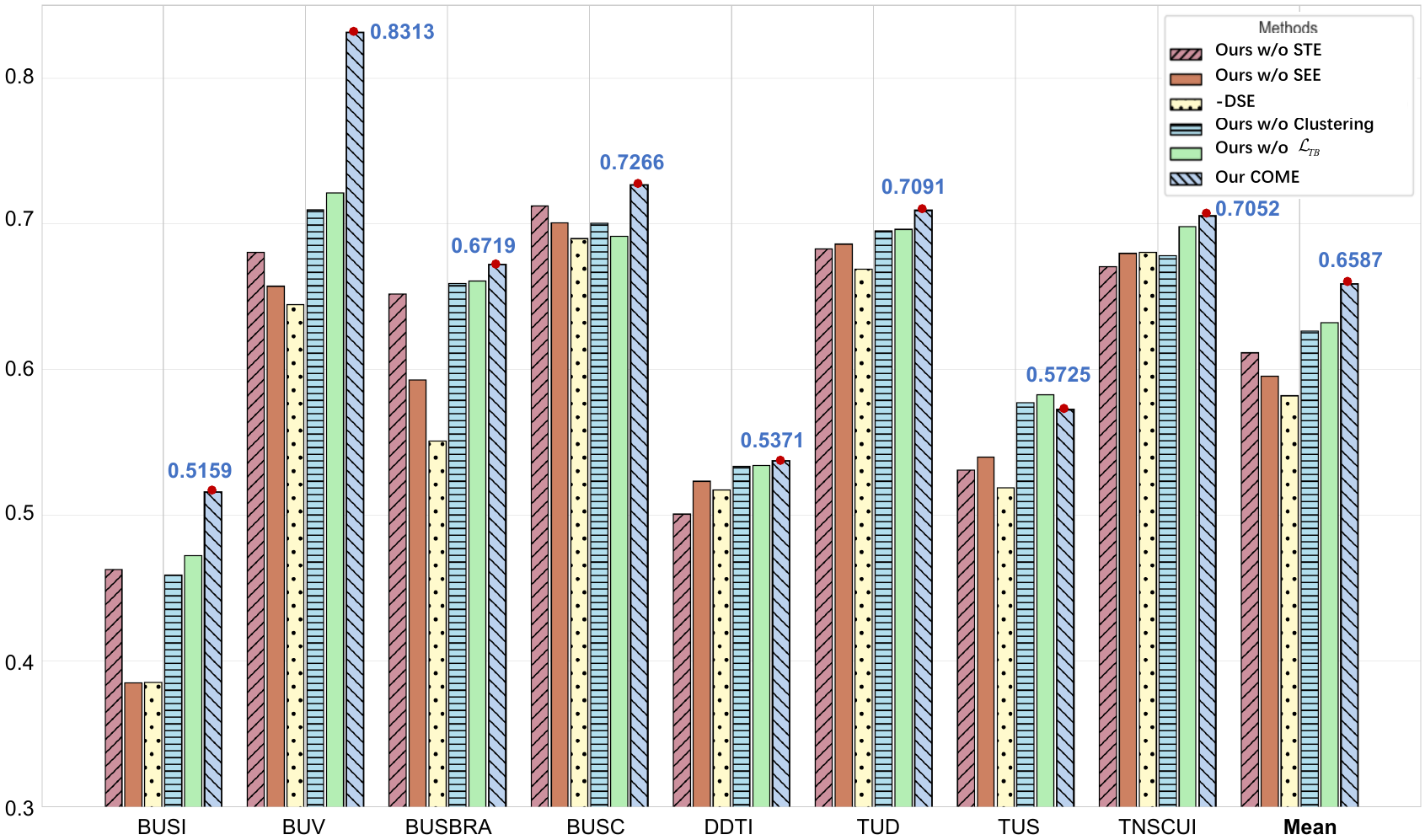}
   \caption{Ablation study on shared experts (i.e., w/o STE, w/o SEE, -DSE), clustering strategies and loss. Sharpest performance drop upon DSE removal validates COME's shared feature space enabling collaborative expert specialization.}
   \label{fig:ablations}
   \vspace{-0.4cm}
\end{figure*}

\textbf{Benchmark Datasets.} We evaluate the proposed COME under three training paradigms for the detection tasks: single dataset, intra-organ consolidation dataset and inter-organ integration dataset. This includes four breast US datasets (BUSI \footnote{https://www.kaggle.com/datasets/sabahesaraki/breast-ultrasound-images-dataset/data.}, BUV \cite{lin2022new}, BUSBRA \cite{gomez2024bus}, BUSC\cite{rodrigues2017breast}) and four thyroid datasets (DDTI \footnote{https://www.kaggle.com/datasets/dasmehdixtr/ddti-thyroid-ultrasound-images.}, TUD \footnote{https://github.com/NEU-LX/TUD-Datebase.}, TUS \footnote{https://www.kaggle.com/datasets/eiraoi/thyroidultrasound.}, TNSCUI \footnote{https://tn-scui2020.grand-challenge.org/.}), as detailed in Figure 1 of the Appendix. All datasets exhibit heterogeneous characteristics from different imaging devices or protocols, manifesting as variations in shadow artifacts, speckle noise patterns, intensity distributions, and the scales of regions of interest.

\noindent\textbf{Implementation Details.} Our framework employs DINO-ViT \cite{zhang2022dino} with an ImageNet-pretrained ResNet-50 backbone and 6-layer transformer. COME modifies the decoder by integrating Mixture-of-Experts (MoE) layers via TUTEL, maintaining core hyper-parameters from \cite{jain2023damex}. We set the capacity factor $f$ to 1.25 and the weight of the ${{\mathcal{L}}_{balance}}$ to 0.1. The number of experts and the Top-K selection settings are discussed in the parameter analysis. The model supports inputs of arbitrary size, trained with a batch size of 8 using the AdamW optimizer (initial learning rate of 1.4e-4). Experiments are conducted on two NVIDIA RTX 4090 GPUs with 24GB of memory. For the clustering operation proposed in COME, we describe it from the following two aspects: i) Fine2Coarse balances global and local features by first clustering into 16 fine centers and then reclustering into 8 coarse centers. ii) Multi-Step Clustering improves label accuracy over five iterations (k=4), using parameters based on empirical results. Additionally, Router Network is a gating mechanism using linear layers from TUTEL \cite{hwang2023tutel}. Datasets are split 8:2 for training/testing. We adopt three training paradigms with mAP computed per dataset: separate training on eight datasets; intra-organ training on mixed thyroid/breast data, then testing on the mixed testset; inter-organ training on all datasets.

\noindent\textbf{Baselines and Metrics.} To validate the effectiveness of the proposed framework, we benchmark COME against SOTA methods in two aspects: 1) Detection Performance and Architectural Benefits: Comparative experiments with representative detectors (DINO \cite{zhang2022dino}, Open-GroundingDINO \cite{liu2024grounding}, YOLOv10 \cite{wang2025yolov10}) and MoE variants (DINO-MoE, DAMEX\cite{jain2023damex}) under single-dataset training to isolate architectural benefits; 2) Universal Capability Evaluation: Comparison with MoE-based architectures and the recently released universal CerberusDet \cite{tolstykh2024cerberusdet} on intra-/inter-organ integration datasets. The quantitative evaluations are based on mean Average Precision (AP) scores.

\subsection{Overall Performance}

\cref{table.2} compares COME with baselines across three training configurations: single-dataset (top), intra-organ (breast/thyroid) combinations (middle), and inter-organ integration (bottom), revealing key findings: 1) DINO (top section) achieves performance with -3.49\% and -4.62\% mean AP decrease over middle/bottom configurations respectively. Besides, the high computational cost of single-dataset training outweighs its performance gains, with poor generalizability causing significant performance drops on unseen data. 2) COME with Multi-Step clustering achieves the highest mean AP (5.14-7.42\% improvement over baselines), demonstrating consistent gains across datasets through multi-source integration and cross-dataset correlation mining. 3) The 1.13\% performance gain between bottom/middle tiers highlights enhanced inter-organ synergy and clinical viability. Notably, our framework achieves 9.79\% (BUSI) and 15.59\% (BUV) superiority over DINO, confirming its viability as a single-source training alternative. Further, We present visualization results of four models trained on the integrated intra-organ benchmark, highlighting image differences and showcasing the superiority of our COME, as shown in the \cref{fig:detection}.

\subsection{Ablation Study}

To validate the effectiveness of each component in COME, we implement five variants: (1) Ours w/o STE; (2) Ours w/o SEE; (3) Ours w/o Dual Shared Experts (-DSE): Discarding dual shared expert modules providing structural-semantic priors; (4) Ours w/o clustering; (5) Ours w/o Traceability loss, as shown in \cref{fig:ablations}. In general, the -DSE exhibits the most severe performance degradation (-7.68\% mean AP), followed by STE and SEE removals. It proves that constructing a universal latent feature space with structural-semantic priors enhances the performance of source-specific experts. Additionally, disabling the clustering (designed for multi-dataset noise suppression) results in 3.24\% scores degradation, while removing the traceability loss (critical for expert specialization and training stability) causes 2.68\% performance decline. In summary, our COME demonstrates superior improvement over baselines on multiple heterogeneous datasets when combining all components. The detailed quantitative results are presented in Appendix Table 1.

\begin{table*}
\centering
\caption{Exploration of the number of experts and Top-K for eight inter-organ integration datasets. Among them, \textbf{bold} indicate optimal performance.}
\label{table:3}
\resizebox{0.9\linewidth}{!}{
\begin{tabular}{c||ccccccccc} 
\toprule
Experts    & BUSI            & BUV             & BUSBRA          & BUSC            & DDTI            & TUD             & TUS             & TNSCUI          & Mean             \\ 
\hhline{=::=========}
\# Experts=4  & 0.4693          & 0.7585          & 0.6811          & \textbf{0.7179} & 0.5137          & \textbf{0.7139} & 0.5880          & 0.6906          & 0.6416           \\
\# Experts=8  & 0.4898          & \textbf{0.8195} & \textbf{0.6919} & 0.6998          & \textbf{0.5508} & 0.7062          & \textbf{0.5883} & \textbf{0.7057} & \textbf{0.6565}  \\
\# Experts=10 & \textbf{0.4932} & 0.8061          & 0.6816          & 0.6953          & 0.5460          & 0.7120          & 0.5619          & 0.6713          & 0.6459           \\ 
\toprule
Top-K    & BUSI            & BUV             & BUSBRA          & BUSC            & DDTI            & TUD             & TUS             & TNSCUI          & Mean             \\ 
\hhline{=::=========}
K=1       & 0.4898          & 0.8195          & \textbf{0.6919} & 0.6998          & \textbf{0.5508} & 0.7062          & \textbf{0.5883} & \textbf{0.7057} & 0.6565           \\
K=2       & 0.4889          & 0.7905          & 0.6852          & 0.7248          & 0.5152          & 0.7067          & 0.5703          & 0.6880          & 0.6462           \\
K=3       & \textbf{0.5159} & \textbf{0.8313} & 0.6719          & \textbf{0.7266} & 0.5371          & 0.7091          & 0.5725          & 0.7052          & \textbf{0.6587}  \\
K=4        & 0.5119          & 0.7646          & 0.6732          & 0.7231          & 0.5283          & \textbf{0.7108} & 0.4762          & 0.6783          & 0.6333           \\
\toprule
\end{tabular}}
\vspace{-0.4cm}
\end{table*}

\begin{table}
\centering
\caption{Domain generalization ability of baselines and COME. By learning shared structure-semantic knowledge and dataset-specific features, COME effectively adapts to other three datasets compared to non-MoE and MoE baseline.}
\label{table:4}
\resizebox{0.9\linewidth}{!}{
\begin{tabular}{c||cccc} 
\toprule
\begin{tabular}[c]{@{}c@{}}Method\\(Train on BUSI)\end{tabular} & BUSI           & BUV            & BUSBRA         & BUSC                     \\ 
\toprule
DINO                                                            & 0.418          & 0.481          & 0.357          & 0.043                    \\
DAMEX                                                           & 0.327          & 0.409          & 0.344          & 0.115                    \\
\begin{tabular}[c]{@{}c@{}}COME \\(Fine2Coarse)\end{tabular}    & \textbf{0.508} & \textbf{0.561} & \textbf{0.410} & \textbf{0.297} \\
\toprule
\end{tabular}}
\vspace{-0.5cm}
\end{table}

\subsection{Parameter Setting}

This section establishes parameter analyses: [1] Configuration of number of source-specific experts $\#Expert \in \{4, 8, 10\}$; [2] Analyzes the impact of $K \in \{1, 2, 3, 4\}$ on the Top‑K selection strategy. This pursuit of optimal parameters is intended to prevent the risk of overloading a single expert while ensuring the optimality of routing decisions through comprehensive parameter space exploration.

\cref{table:3} shows the quantitative results, and we make the following observations: 1) COME achieves the best mean performance, peaking at 8 experts. In particular, the increase is prominent in BUV (\underline{\textbf{1.34-6.1\%}}). It is worth noting that as \#Experts aligns with dataset scale ($N=8$ for 8 training sources), peak performance forms a theoretical closed-loop with the traceability loss design, validating "single-expert specialization in source-specific feature extraction." 2) $K=3$ yields optimal performance (\underline{\textbf{0.22-2.54\%}} gains), showing that moderate expert activation reduces sparsity, maintains cross-source generalization through a shared feature space, and enhances representation by combining complementary expert knowledge.

\subsection{In-Depth Analysis}

We further highlight the advantages of the proposed COME by answering the following questions.

\noindent\textbf{Q1: Does COME demonstrate domain generalization capability?} 

Four breast datasets exhibit significant distribution variations, such as source-specific noise and artifacts. To validate COME's generalizability, we randomly train on the BUSI and test directly on the other three breast datasets.

The performance comparison of DINO (non-MoE setup), DAMEX and our COME is demonstrated in \cref{table:4}. Experimental results show significant performance gains: COME achieves a 9\% mean AP improvement on BUSI and an aggregate +32.8\% mean AP increase across three additional datasets. This proves that our dual structure-semantic shared experts effectively provide universal knowledge while the S²E module successfully captures dataset-specific representations through feature routing.

\noindent\textbf{Q2: How do feature representations evolve across clustering stages?} 

\begin{figure}[!t]
  \centering
  \begin{minipage}{\linewidth}
    \centering
    \includegraphics[width=0.99\linewidth, height=2.4cm]{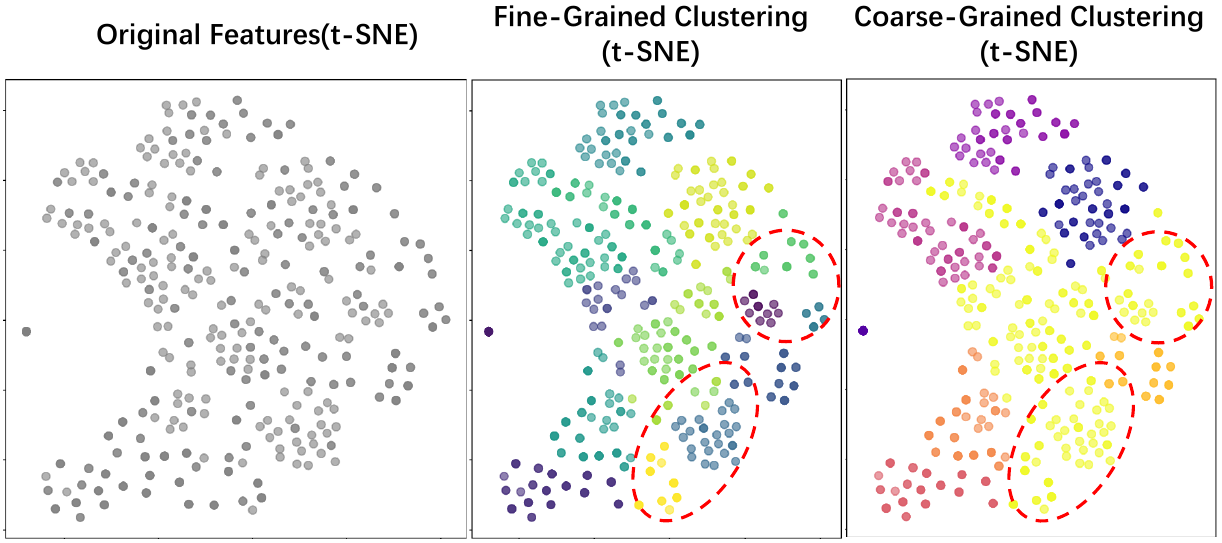}
    \subcaption{Visualization of Fine2Coarse clustering.}
  \end{minipage}
  
  \vspace{0.2cm}  
  
  \begin{minipage}{\linewidth}
    \centering
    \includegraphics[width=0.7\linewidth]{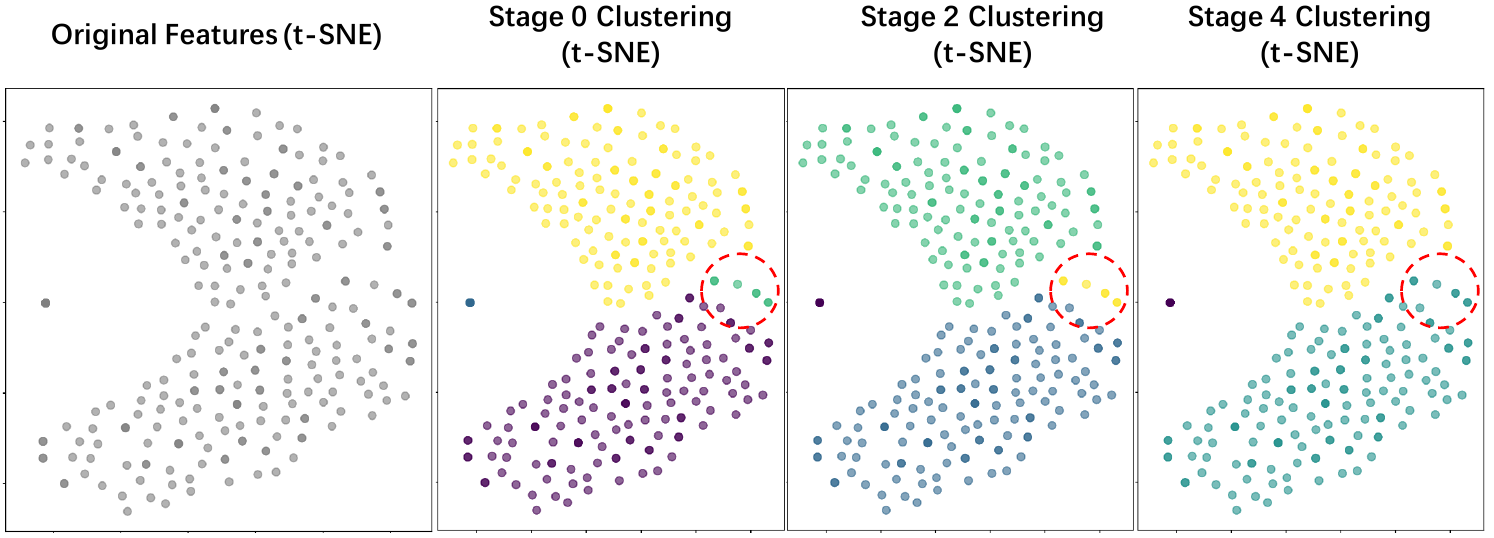}  
    \subcaption{Visualization of Multi-Step clustering.}
  \end{minipage}
  
  \caption{The upper image (a) and the lower image (b) represent the feature visualization using the Fine2Coarse and Multi-Step clustering strategies, respectively. As the clustering stage progresses, the clustering results improve significantly (as indicated by the red circle).}
  \label{fig:clustering}
\end{figure}

The proposed COME achieves optimal performance with Fine2Coarse clustering strategy. \cref{fig:clustering}(a) illustrates three-stage feature distributions: initial/fine-grained (16-center)/coarse-grained (8-center) clustering phases. Fine-grained clustering introduces localized fragmentation that affects S²E training stability, while coarse-grained clustering enhances feature concentration by routing semantically similar samples to unified experts for enhanced specialization. Simultaneously, we demonstrate a Multi-Step clustering in \cref{fig:clustering}(b): We implement a five-step clustering, visualizing three of these steps. The clustering results from the previous stage serve as priors for the current stage, enabling COME to progressively construct independent attribute subspaces and achieve finer, more diversified feature representations that enhance downstream inference tasks.

\noindent\textbf{Q3: How does Recall perform on imbalanced and heterogeneous data?}

We adopt mean mAP to evaluate COME, as it's widely used in detection models across diverse datasets \cite{wang2024uni, lei2024exploring}. Afterwards, we further include Recall as the evaluation metric in Table \ref{tab:R1}, with COME remaining optimal.

\begin{table}[h]
\centering
\caption{Comparison based on Recall.}
\label{tab:R1}
\resizebox{0.47\textwidth}{!}{
\begin{tblr}{
  column{odd} = {c},
  column{2} = {c},
  column{4} = {c},
  vline{2-6} = {-}{0.08em},
  hline{1-2,6} = {-}{0.08em},
}
Method                 & BUSI~ & BUV~ & BUSBRA & BUSC~ & Mean \\
DINO              & 0.43  & 0.58 & 0.62   & 0.54  & 0.54 \\
DINO-MoE          & 0.50  & 0.75 & 0.67   & 0.55  & 0.62 \\
DAMEX             & 0.45  & 0.77 & 0.68   & 0.69  & 0.65 \\
COME(Fine2Coarse) & 0.55  & 0.80 & 0.70   & 0.70  & \textbf{0.69} 
\end{tblr}}
\end{table}

\noindent\textbf{Q4: How does COME choose between Fine2Coarse and Multi-Step Clustering strategies?}

Firstly, two clustering strategies are independent: Fine2Coarse balances local and global features by avoiding feature fragmentation and preserving detail, while Multi-Step Clustering improves semantic continuity and reduces heterogeneity. To evaluate the potential for practical employment, we apply both strategies together. As shown in Table~\ref{tab:R2}, the mixing approach acquires comparable performance on intra-breast dataset but incurs higher inference time. Overall, given the comparable mean AP and faster inference to Multi-Step, Fine2Coarse is adopted as the default in COME, with Multi-Step serving as a optional alternative under ample computational resources.

\begin{table}[h]
\centering
\caption{Comparison of Clustering Operations in COME.}
\label{tab:R2}
\resizebox{0.46\textwidth}{!}{
\begin{tabular}{c!{\vrule width \heavyrulewidth}c!{\vrule width \heavyrulewidth}c!{\vrule width \heavyrulewidth}c!{\vrule width \heavyrulewidth}c} 
\toprule
Clustering    & Intra-breast & Intra-thyroid & Inter-organ & \multicolumn{1}{l}{Time} ($\mu s$)  \\ 
\toprule
Multi-Step~ ~ & 0.6626               & 0.6322                & 0.6558              & 615                      \\
Fine2Coarse   & 0.6471               & 0.6344                & 0.6587              & \textbf{132}                      \\
Mixing        & 0.6533               & 0.6140                & 0.6337              & 845                      \\
\toprule
\end{tabular}}
\end{table}




\section{Conclusion}
\label{conclusion}

This study proposes a dual structure-semantic learning framework for multiple heterogeneous US analysis, demonstrating superior performance in lesion detection. Addressing multi-source data characteristics, we innovatively design two shared experts to construct a universal feature space by integrating structural consistency and semantic invariance, supporting source-specific feature mining. Then a source-specific expert routing module employs clustering for feature decoupling, enhanced by traceability loss to reinforce expert specialization and ensure consistent data routing. Additionally, empirical results validate the effectiveness of intra-/inter-organ data integration strategies, offering new perspectives for the US analysis community, particularly given the increasing availability of public datasets.

In the future, we plan to follow the advancements of cutting-edge technologies (e.g. MoE) to continuously optimize our model, expanding its applicability to encompass a broader array of dataset domains and a wider range of medical tasks.

\section{Acknowledgments}
This work was supported in part by National Nature Science Foundation of China grants (62271246); the Science and Technology Commission of Shanghai Municipality (Nos.24511104100).

{\small
\bibliographystyle{ieee_fullname}
\bibliography{main}
}

\newpage
\clearpage

\section{Overview}
\begin{itemize}
    \item Limitations (\S \ref{sec:limit})
    \item Details of Dataset (\S \ref{sec:annotation})
    \item Additional Qualitative Evaluation (\S \ref{sec:case})
    \item Details of Ablations (\S \ref{sec:ablations})
    \item Addition Parameter Setting on Intra-organ Datasets (\S \ref{sec:parameter})
    \item Feature Visualization during Clustering  (\S \ref{sec:clustering})
\end{itemize}

\section{Limitations} \label{sec:limit}
We must also candidly acknowledge some limitations in our research, specifically: 
1) Model: Since the MoE model is still under development, the current design is not yet cutting-edge. For instance, techniques like using LoRA for weight integration and training larger MoE models have not been fully implemented.
2) Application scenarios: Although our study has encompassed nearly all mainstream datasets, there is potential for further extension to other scenarios to verify generalizability.
3) Types of tasks: Our COME focuses on multi-source heterogeneous datasets. The next step is to expand its capabilities to comprehensively address a wider range of medical tasks.

\section{More Details} 
\subsection{Details of Dataset} \label{sec:annotation}


To develop a universal model for heterogeneous ultrasound (US) datasets, we built a benchmark of 4 breast and 4 thyroid US datasets. These datasets come from different sources and exhibit significant domain differences, such as variations in shadow artifacts, speckle noise, grayscale levels, and anatomical structures, as shown in \cref{fig:datasetA1}.

\begin{figure}[h]
    \centering
    \includegraphics[width=0.8\linewidth]{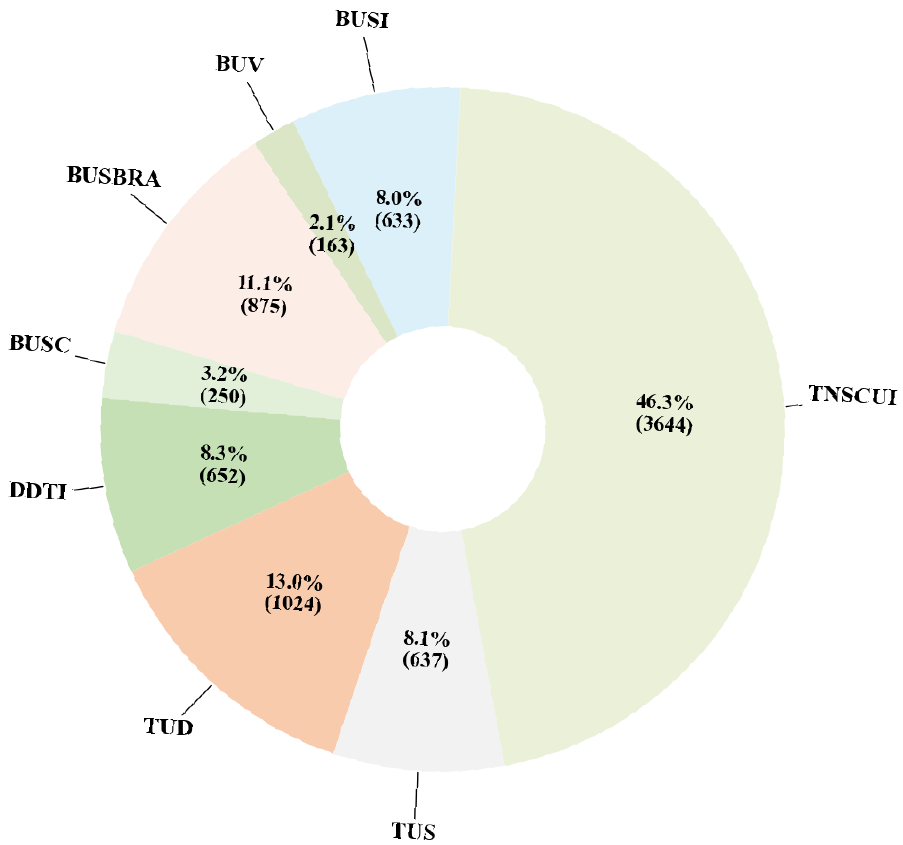}
    \caption{The dataset proportions and corresponding image counts in the benchmark.}
    \label{fig:dataset2}
\end{figure}

Due to strict collection conditions and reliance on expert doctors, many US datasets are imbalanced (see \cref{fig:dataset2}). However, the proposed COME architecture effectively overcomes this issue, as demonstrated in Table 2 of the main text.

\subsection{Additional Qualitative Evaluation} \label{sec:case}
In the main text, we select one sample per dataset for comparison. Here, \cref{fig:comprison_result} shows additional lesion detection examples, demonstrating that our structure-semantic learning-based COME model delivers robust performance on diverse US images and holds promise for real-world clinical applications.

\begin{table}[t]
\renewcommand{\thetable}{2}
\centering
\caption{Effect of the number of experts on the intra-organ (thyroid) integration dataset.}
\label{table.2}
\begin{tabular}{@{}c||cccc|c@{}}
\toprule
\# Experts & TUD & TUS & DDTI & TNSCUI & Mean \\
\toprule
2  & 0.5095 & \textbf{0.7092} & \textbf{0.5794} & 0.6921 & 0.6226 \\
4  & 0.5594 & 0.6932 & 0.5750 & \textbf{0.6935} & 0.6303 \\
8  & 0.5481 & 0.7032  & 0.5900 & 0.6875 & \textbf{0.6322} \\
10 & \textbf{0.5595}  & 0.6919 & 0.5580 & 0.6878 & 0.6243 \\
\toprule
\end{tabular}
\end{table}

\subsection{Details of Ablations} \label{sec:ablations}
In the main text, Figure 4 shows qualitative ablation visualizations. Here, \cref{table.1} provides quantitative results that further demonstrate the effectiveness of each COME component.

\begin{table*}
\renewcommand{\thetable}{1}
\belowrulesep=0pt
\aboverulesep=0pt
\centering
\refstepcounter{table}
\label{table.1}
\caption{Quantitative Performance of the ablation study.}
\resizebox{\textwidth}{!}{
\begin{tabular}{l||cccccccc||c}
\toprule
Method  & BUSI   & BUV    & BUSBRA & BUSC   & DDTI   & TUD    & TUS    & TNSCUI & Mean    \\ 
\hhline{=::========::=}
STE    & 0.4628 & 0.6802 & 0.6517 & 0.7123  & 0.5008 & 0.6827 & 0.5307 & 0.6704 & 0.6115  \\
SEE                        & 0.3849 & 0.6570  & 0.5927 & 0.7006 & 0.5231 & 0.6859 & 0.5397 & 0.6795 & 0.5954   \\
Dual Shared Experts(-DSE) & 0.3853 & 0.6445 & 0.5509 &0.6897 & 0.5173 & 0.6687 & 0.5189 & 0.6802 & 0.5819   \\
Clustering                 & 0.4587 &0.7093& 0.6590  & 0.7003 & 0.5335 & 0.6952 & 0.5772 & 0.6779 & 0.6264   \\
Traceability Loss          & 0.4721 & 0.7211& 0.6605 & 0.6913 & 0.5341 & 0.6960  & {\cellcolor[rgb]{0.859,0.937,0.98}}\textbf{0.5826} & 0.6981 & 0.6320   \\
\rowcolor[rgb]{0.824,0.827,0.827} Our COME(Fine2Coarse)    & {\cellcolor[rgb]{0.859,0.937,0.98}}\textbf{0.5159} & {\cellcolor[rgb]{0.859,0.937,0.98}}\textbf{0.8313} & {\cellcolor[rgb]{0.859,0.937,0.98}}\textbf{0.6719} & {\cellcolor[rgb]{0.859,0.937,0.98}}\textbf{0.7266} & {\cellcolor[rgb]{0.859,0.937,0.98}}\textbf{0.5371} & {\cellcolor[rgb]{0.859,0.937,0.98}}\textbf{0.7091} & 0.5725 & {\cellcolor[rgb]{0.859,0.937,0.98}}\textbf{0.7052} & {\cellcolor[rgb]{0.859,0.937,0.98}}\textbf{0.6587}   \\
\bottomrule
\end{tabular}}
\end{table*}

\subsection{Parameter Setting on Intra-organ Datasets} \label{sec:parameter}


In the main text, we explore how the number of experts affects COME's performance on the inter-organ integrated dataset. Here, we evaluate its sensitivity on the intra-organ thyroid dataset (see \cref{table.2}). dataset-specific experts through traceability loss and heterogeneous architecture, effectively isolating source features and minimizing interference. To ensure comprehensive analysis, we specifically include a 2-expert configuration for the intra-organ dataset with four sources, demonstrating its viability under constrained conditions.


\section{Feature Visualization during Clustering} \label{sec:clustering}
\begin{figure*}[h]
    \centering
    \includegraphics[width=0.8\linewidth]{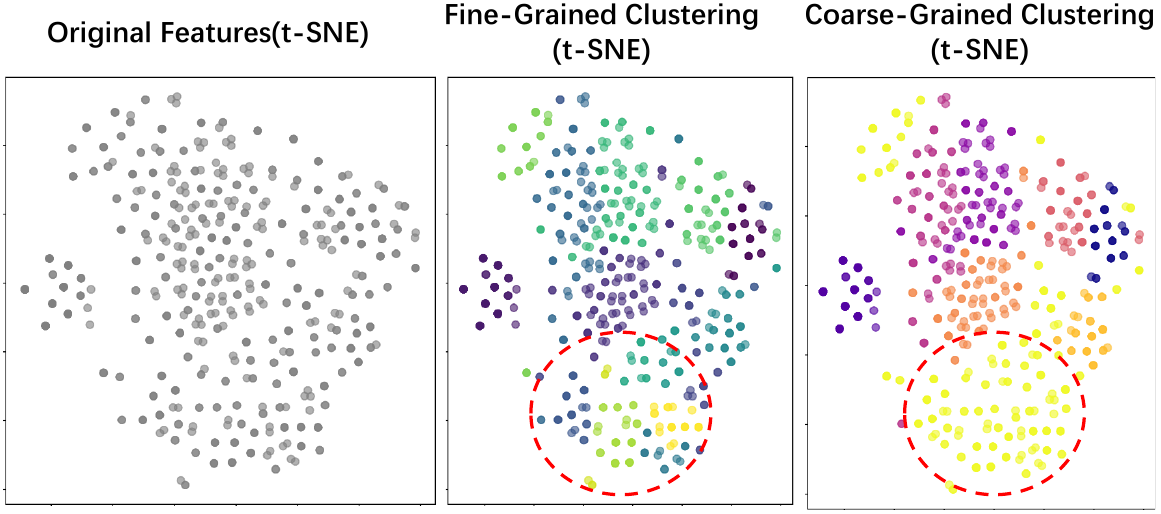}
    \caption{Feature visualization during training of the FineCoarse hierarchy clustering.} 
    \label{fig:clustering1}
\end{figure*}

The proposed COME achieved optimal performance with its Fine2Coarse clustering strategy. \cref{fig:clustering1} presents additional samples illustrating the feature distributions.

\begin{figure*}[h]
    \centering
    \includegraphics[width=0.8\linewidth]{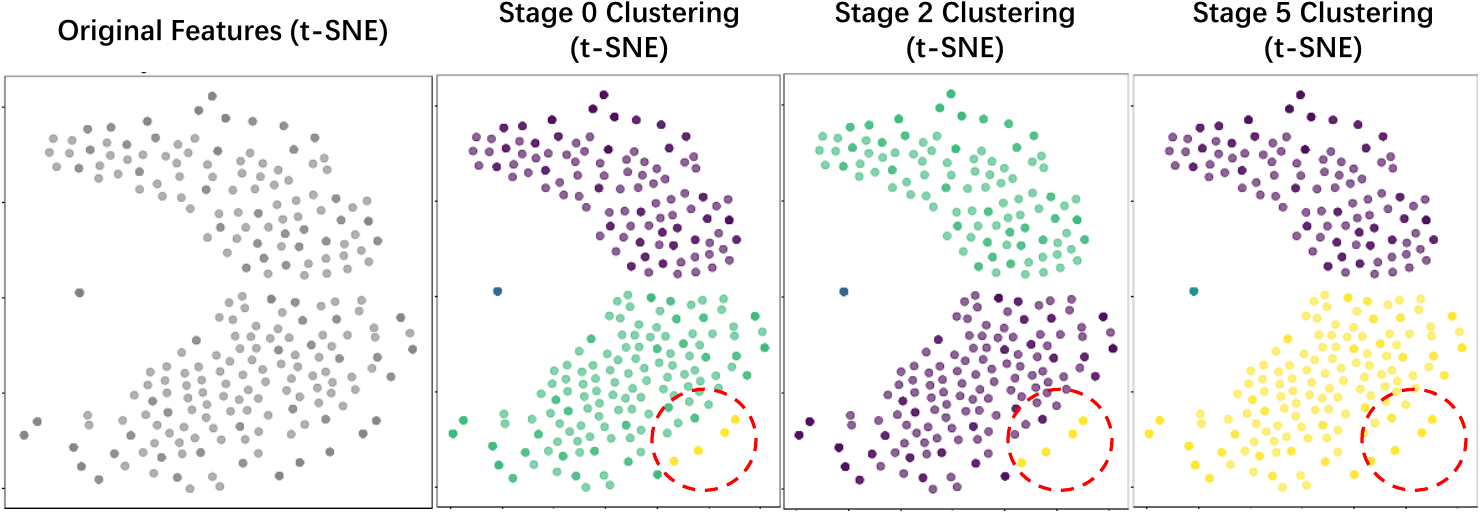}
    \caption{Feature visualization during training of COME's Multi-Step clustering.}
    \label{fig:clustering2}
\end{figure*}

\begin{figure*}[h]
    \centering
    \includegraphics[width=0.8\linewidth]{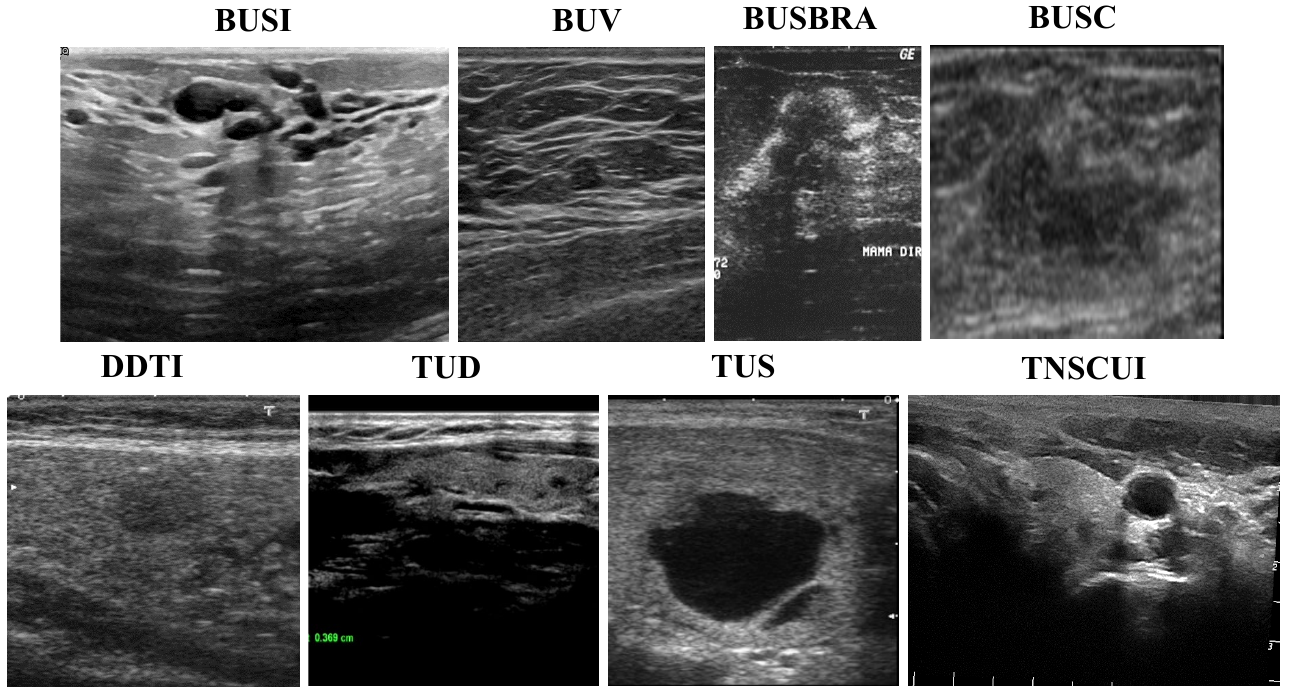}
    \caption{This paper constructs an integrated benchmark comprising eight heterogeneous breast and thyroid US datasets. And the distinct characteristics of each dataset pose challenges in building a universal analysis framework.}
    \label{fig:datasetA1}
\end{figure*}

\begin{figure*}[h]
    \centering
    \includegraphics[width=\linewidth]{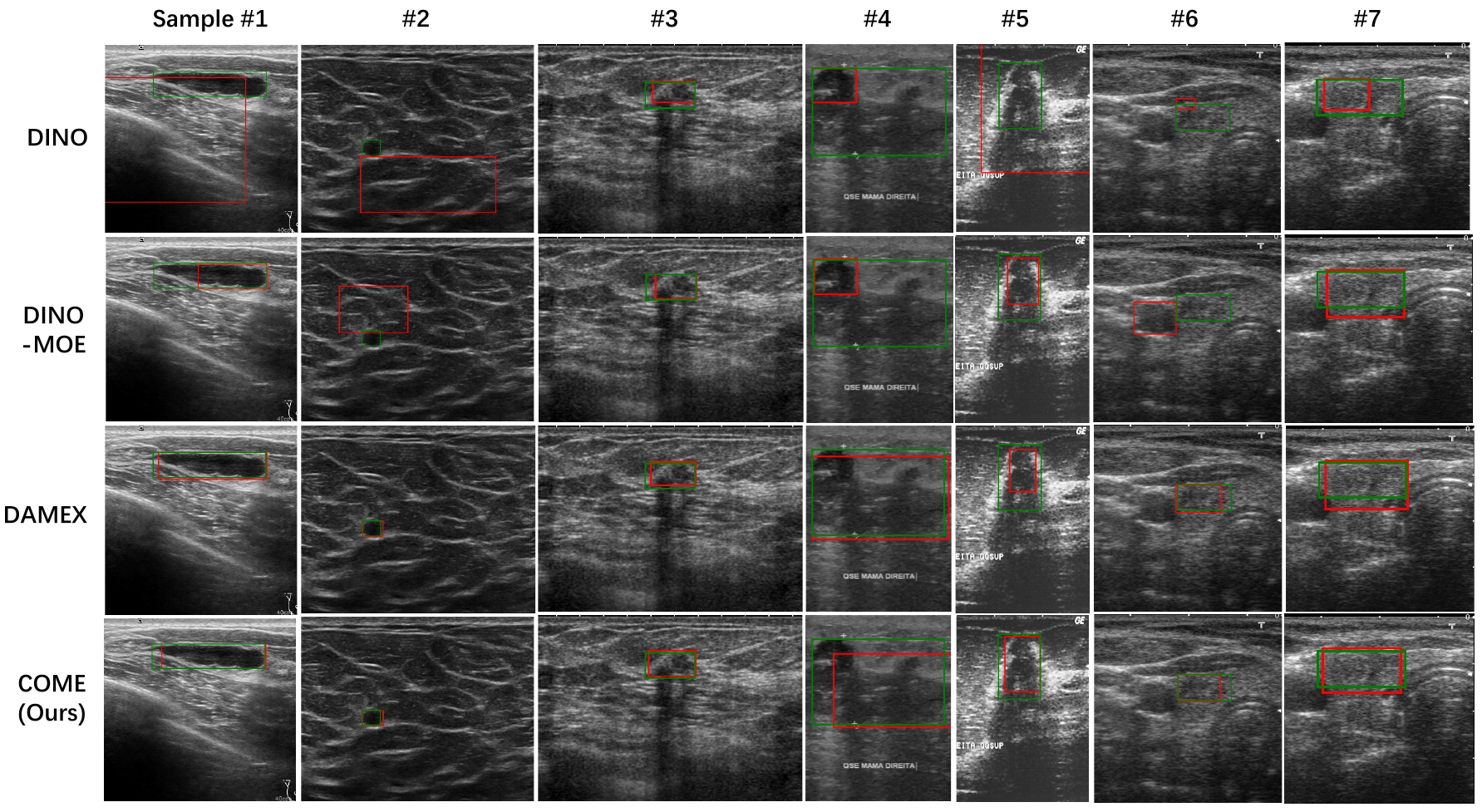}
    \caption{Additional lesion detection examples from the inter-organ integrated dataset.}
    \label{fig:comprison_result} 
\end{figure*}

Simultaneously, we demonstrate a multi-step clustering in \cref{fig:clustering2}.

\end{document}